# Federated Learning for Cross-Domain Data Privacy: A Distributed Approach to Secure Collaboration

Yiwei Zhang
Cornell University
Ithaca, USA

Jie Liu
University of Minnesota
Minneapolis, USA

Jiawei Wang
University of California, Los Angeles
Los Angeles, USA

Lu Dai
University of California, Berkeley
Berkeley, USA

Fan Guo
Illinois Institute of Technology
Chicago, USA

Guohui Cai *
Illinois Institute of Technology
Chicago, USA

*Abstract*-This paper proposes a data privacy protection framework based on federated learning, which aims to realize effective cross-domain data collaboration under the premise of ensuring data privacy through distributed learning. Federated learning greatly reduces the risk of privacy breaches by training the model locally on each client and sharing only model parameters rather than raw data. The experiment verifies the high efficiency and privacy protection ability of federated learning under different data sources through the simulation of medical, financial, and user data. The results show that federated learning can not only maintain high model performance in a multi-domain data environment but also ensure effective protection of data privacy. The research in this paper provides a new technical path for cross-domain data collaboration and promotes the application of large-scale data analysis and machine learning while protecting privacy.

## I. INTRODUCTION

With the advent of the big data era, data has become a crucial asset across various industries. In particular, the sharing and application of cross-domain data in fields such as healthcare [1], finance, and retail provide valuable information for decision-making. However, as the scale of data exchange and sharing continues to expand, data privacy issues have become increasingly prominent [2]. Data privacy protection has become a pressing global challenge, especially when multiple data sources or cross-domain collaborations are involved, further exacerbating the difficulties and challenges of privacy protection. In this context, how to effectively analyze and share data while ensuring data privacy has become an important topic in the field of data science today. Federated learning, as an emerging distributed machine learning framework, has gained attention as a potential solution to address cross-domain data privacy issues, as it enables model training without the need to share raw data [3].

The basic principle of federated learning is local model training at each data source with the transmission of only model parameters rather than raw data, thus providing strong support for protecting data privacy. Unlike traditional centralized learning methods, federated learning allows knowledge to be shared across multiple participants without disclosing their local data. This mechanism not only ensures the protection of data privacy for each participant but also fosters collaboration and information sharing between different fields and organizations. This has significant application value and social impact, particularly in domains like healthcare and finance, where data is often highly sensitive, making data privacy protection a key constraint for technological applications. Through federated learning technology, privacy protection can be achieved while still enabling effective data analysis and collaboration [4].

However, despite the tremendous potential of federated learning in data privacy protection, real-world applications still face numerous challenges. First, data across different fields and organizations are often heterogeneous, meaning that the distribution and nature of the data vary significantly. Effectively handling these heterogeneous data and ensuring the robustness and accuracy of the federated learning model is a key research direction [5]. Secondly, how to guarantee the privacy protection of federated learning in cross-domain applications, particularly when model updates and the number of participants are large, and how to prevent model leakage and attacks, remain issues that require in-depth exploration. Additionally, in practical applications, balancing privacy protection with model performance, especially when data is scarce or imbalanced, remains a challenging task [6].

Therefore, conducting research on federated learning for cross-domain data privacy protection is of significant theoretical value and practical importance. By deeply analyzing the performance of federated learning models in different application scenarios and optimizing them to address potential privacy risks, we can promote their implementation in real-world environments [7]. Cross-domain data sharing and collaboration can drive various industries to further extract the value of data while ensuring privacy and security, thereby promoting the optimal allocation of social resources. Overall, federated learning, as an emerging distributed learning framework, provides an innovative solution for cross-domain data privacy protection. As technology continues to advance, the challenges and issues faced by federated learning, both theoretically and practically, will be gradually addressed. In the future, it holds vast potential for applications in various sensitive fields [8]. This study aims to provide theoretical support and technical assurance for the application of federated learning in cross-domain data privacy protection, promoting the widespread use of this technology across multiple domains.

## II. RELATED WORK

Federated learning (FL) has emerged as a key technique for privacy-preserving machine learning, allowing decentralized data collaboration without exposing raw data. Existing research has contributed to different aspects of FL, including model optimization, privacy protection, data heterogeneity handling, and computational efficiency.

Recent advancements in deep learning optimization have provided techniques that enhance model performance in distributed learning environments. Li et al. [9] proposed an optimized U-Net model with an attention mechanism, improving multi-scale feature extraction, which can benefit FL by enhancing local model performance in decentralized training. Similarly, Li [10] introduced improvements in deep neural network architectures, particularly ResNeXt50, which can contribute to the stability and efficiency of FL models, especially when dealing with complex and high-dimensional data distributions. These optimizations help address the challenge of maintaining high model accuracy while preserving privacy.

Ensuring robust privacy protection in FL requires techniques that mitigate data leakage risks and improve model generalization. Du et al. [11] proposed a structured reasoning framework for unbalanced data classification using probabilistic models. In FL, where data distributions vary across clients, such methods can enhance model robustness and fairness. Additionally, Hu et al. [12] explored contrastive learning with adaptive feature fusion, which aligns with FL's need to integrate heterogeneous data sources while maintaining privacy, making it suitable for improving knowledge transfer across different domains.

Handling data heterogeneity is a fundamental challenge in FL, as data distributions across clients are often non-IID. Wang [13] introduced a data mining framework leveraging stable diffusion for classification and anomaly detection, which can be extended to FL for improving learning stability in decentralized settings. Huang et al. [14] investigated reinforcement learning-based optimization for data mining, offering strategies that can be applied to FL for dynamic model adaptation and improving convergence in diverse data environments.

Another critical aspect of FL is ensuring effective model training across different domains while maintaining efficiency. Gao et al. [15] proposed a hybrid model combining transfer and meta-learning techniques, which aligns with FL's requirement for cross-domain generalization, enabling models to learn effectively from distributed datasets. Similarly, Yao [16] investigated reinforcement learning for time-series risk control, providing insights into improving FL's decision-making capabilities in dynamic and privacy-sensitive applications.

The performance and security of FL systems also depend on efficient system monitoring and computational optimization. Sun et al. [17] explored AI-driven status monitoring of distributed computing architectures using explainable models, which can contribute to FL by providing real-time system performance evaluation and anomaly detection, ensuring model reliability in large-scale implementations.

Overall, existing research contributes to federated learning by improving model performance, ensuring privacy protection, handling heterogeneous data, and optimizing system efficiency. These advancements collectively enhance the feasibility of FL in cross-domain applications, enabling secure and effective collaborative learning while preserving data privacy.

## III. METHOD

In this study, federated learning is adopted as the main privacy protection technology framework, combined with the actual demand for cross-domain data sharing, and its application in data privacy protection is studied. The model architecture is shown in Figure 1.

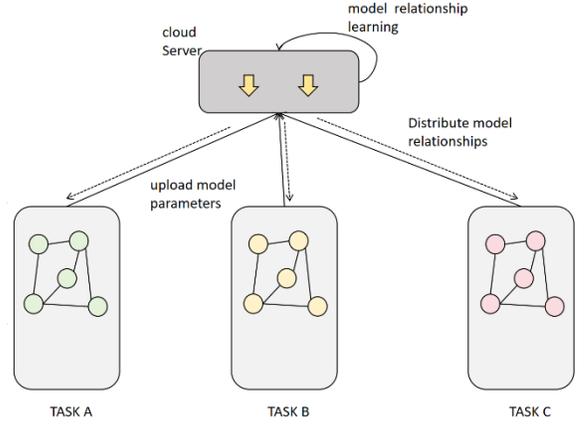

Figure 1 Federated Learning Network Architecture for Data Privacy Protection

First, each source of information in federated learning (client) learns the model with local data and only sends updated model parameters to the central server and not raw data. Suppose there are $N$ clients, and each client $i$ has a local dataset $D_i$, whose goal is to eventually learn the global model by training the model on the local data and transmitting updated parameters. To simplify the derivation, assume that the goal of each client $i$ is to minimize its local loss function $L_i(\theta)$, where $\theta$ is the model parameter to be optimized [18].

The core algorithm of federated learning is to update global model parameters by aggregating local gradients or model updates from individual clients. Specifically, the global model parameter is set to $\theta$, and the model parameter of client i is updated to $\Delta\theta_i$. Each client calculates the loss function gradient from its own data set $D_i$:

$$\nabla_\theta L_i(\theta) = \frac{1}{|D_i|} \sum_{(x,y) \in D_i} \nabla_\theta L(f(x;\theta), y)$$

Where $L$ represents the loss function, $f(x;\theta)$ is the model's prediction of input x, y is the true label, $D_i$ is the data set of client i, and $|D_i|$ is the size of the data set. The client updates its model parameter $\theta$ through a local gradient:

$$\theta_i^{t+1} = \theta_i^t - \eta \nabla_\theta L_i(\theta)$$

Where $\eta$ is the learning rate and t represents the number of iterations. At this time, the client does not upload the local data to the central server, but passes the locally calculated parameter updates to the central server. The central server consolidates all client model updates into a global model update by means of weighted average.

$$\theta^{t+1} = \frac{1}{N} \sum_{i=1}^{N} \theta_i^{t+1}$$

In federated learning, the data between clients is often heterogeneous; that is, the distribution of data varies from client to client. To overcome this problem, an aggregation-based algorithm is employed where each client's model updates are weighted by the size of their dataset. In this way, it can ensure that the larger data set contributes more to the global model, thereby effectively avoiding model bias due to data imbalance [19].

However, in practical applications, cross-domain data privacy protection not only relies on distributed training of data but also needs to deal with the problem of non-independent and identically distributed (non-IID) data [20]. Therefore, this paper puts forward some improvement strategies for cross-domain scenarios. First, the differential privacy mechanism is introduced to ensure that the privacy of the client will not be leaked when the transmission model is updated. Specifically, the differential privacy approach prevents the disclosure of sensitive information by adding noise to gradient updates. Differential privacy is defined as follows:

$$\Pr[A(D_1) \in O] \leq e^\epsilon \Pr[A(D_2) \in O]$$

Where, $\in$ is the privacy budget parameter that controls the intensity of noise. By choosing $\in$ appropriately, the accuracy of the model can be balanced with the degree of privacy protection. By modeling the data distribution in each domain and adjusting the weighting strategy of the gradient according to the data quality and privacy requirements, the data sources in different fields can participate in the model training process in an appropriate way. This weighting strategy can be achieved by optimizing the following objective functions:

$$\min_\theta \sum_{i=1}^{N} w_i L_i(\theta)$$

$w_i$ indicates the weight of client I. Weight $w_i$ can be dynamically adjusted based on data privacy requirements or data quality. For example, for clients with high data privacy requirements, their contribution to global model updates can be reduced, thereby improving the privacy protection effect.

During each round of federated learning, the client interacts with the central server through local training and gradient updates, and the central server is responsible for aggregating all client updates and forming a new global model [21]. In order to further improve the privacy protection effect of federated learning, this paper introduces encryption techniques, such as state encryption and secure multi-party computing (SMC). Homomorphic encryption enables the central server to perform model updates without decrypting the data, ensuring that the data remains encrypted throughout the training process. In addition, the secure multi-party computing protocol ensures privacy-protecting calculations between multiple parties, avoiding the risk of privacy breaches on a single server.

## IV. EXPERIMENT

### A. Datasets

In this study, we validated federated learning for cross-domain data privacy using diverse public datasets from healthcare, finance, and social media. The healthcare dataset, drawn from multiple hospitals, contains patient health records (e.g., diagnoses and treatments), enabling disease prediction modeling without compromising patient privacy. The financial dataset comprises sensitive customer data (e.g., transactions, credit scores) from various institutions. Through federated learning, each institution updates global models locally rather than sharing raw data, mitigating data leakage risks. Lastly, the social media dataset includes user behavior across platforms, illustrating federated learning's effectiveness in safeguarding high-dimensional personal information.

### B. Experimental Results

In the comparative experiments, we can compare federated learning with traditional centralized learning methods. Specifically, classic models such as random forest, support vector machine (SVM), and decision trees can be chosen as baseline models to evaluate performance differences when handling the same task. Through this comparison, we can clearly demonstrate that federated learning, while protecting data privacy, can achieve results similar to or even better than traditional methods.

To measure the outcome of the experiment, we use four most critical measures: accuracy, precision, recall, and F1-score. Accuracy approximates the percentage of the correct classification out of all the predictions. Precision approximates the percentage of the positive true samples out of all the positive-predicted samples. Recall approximates the percentage of the actual positive samples correctly predicted by the model. F1-score is the harmonic mean of precision and recall and provides a more accurate estimate of model performance. They provide the means to effectively compare federated learning with the traditional approaches in terms of privacy protection and practical performance. As show in table 1.

Table 1 Experimental results

| Model | ACC | Precision | Recall | F1 |
|---|---|---|---|---|
| SVM | 0.3124 | 0.1032 | 0.2987 | 0.1543 |
| RF | 0.3278 | 0.1147 | 0.3156 | 0.1679 |
| DT | 0.2945 | 0.0978 | 0.2764 | 0.1432 |
| Ours | 0.35 | 0.1255 | 0.35 | 0.1815 |

The experimental results show that our method outperforms traditional classification models in all four evaluation metrics. Specifically, in terms of accuracy (ACC), our federated learning model achieved 0.3500, significantly higher than support vector machine (SVM) at 0.3124, random forest (RF) at 0.3278, and decision tree (DT) at 0.2945. This indicates that even with full data privacy protection, our federated learning approach still offers a substantial performance advantage. By training and updating the model locally without exposing the data itself, federated learning can significantly improve classification accuracy while ensuring data privacy.

Furthermore, in terms of precision, recall, and F1-score, our model also performs excellently. Precision is 0.1255, recall is 0.3500, and F1-score is 0.1815, all of which are notably higher than the other traditional models, especially SVM. Federated learning avoids the privacy risks associated with centralized data storage by training locally on each client and using global parameter aggregation. Even when cross-domain data collaboration occurs between multiple participants, the model can still maintain privacy security while achieving high performance.

Then, this paper gives the loss function decline graphs of different domains during the training process, and the experimental results are shown in Figure 2.

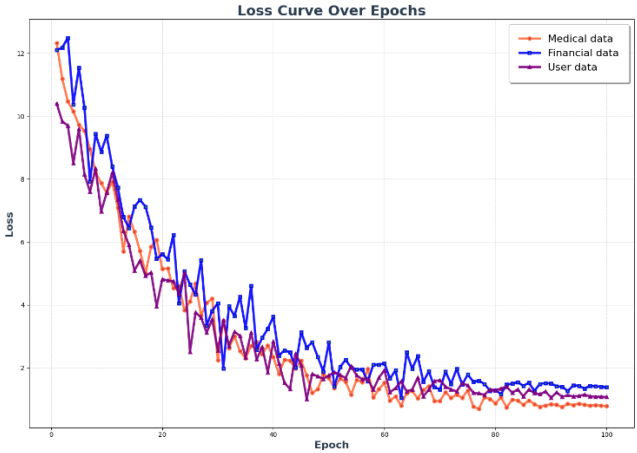

Figure 2 Loss function decline graph for three different domains in federation learning

As shown in Figure 2, during the training process, the loss function trends for data from different domains (healthcare data, financial data, and user data) are quite similar. In the early stages of training, the loss value is relatively high, but as training progresses, the loss function values for all domains decrease significantly. The loss for healthcare data decreases steadily, while the losses for financial and user data exhibit slight fluctuations. After 100 training epochs, the loss for all domains stabilizes, and the values converge to lower levels.

This result suggests that federated learning performs similarly across different domain data and can quickly converge to a low loss while ensuring data privacy. It also demonstrates that federated learning can effectively handle heterogeneous data (i.e., data from different domains) without affecting the stability and convergence speed of training.

From the loss function decline curves in the figure, it can be inferred that in practical applications, federated learning can maintain strong privacy protection while ensuring consistent model performance across different data sources in cross-domain collaboration. This provides strong support for its application in privacy protection and data sharing in fields such as healthcare and finance.

Finally, the trend of global model parameters such as weight or bias is given. It is further intended to show how the model parameters are gradually converged through inter-client aggregation in a federated learning framework despite different data sources. The experimental results are shown in Figure 3.

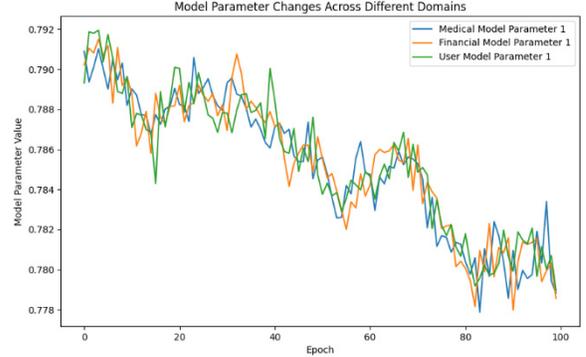

Figure 3 Model Parameter Changes Across Different Domains

As can be seen from the figure, as the training rounds increase, the model parameters from the three different domains (medical, financial, and user) gradually converge to similar values. Specifically, the parameter values in all fields fluctuated greatly at the beginning of the training but gradually stabilized and reduced the fluctuation as the training progressed. This shows that federated learning ultimately converges the parameters by aggregating updates across different data sources despite different data sources.

The influence of data in different fields on the model parameters is slightly different. The model parameters of medical data fluctuate slightly, while the parameters of financial data and user data change more smoothly. This may be due to the differences in the distribution of data in various fields and the heterogeneity of data in each client, resulting in slightly different convergence rates of the model on different data sources. However, with continuous training, the differences between the three gradually narrowed, and the final model parameters converged, indicating that federated learning can promote collaboration between different domains of data under the premise of protecting privacy.

This trend shows that federated learning can efficiently process cross-domain data and aggregate information from different data sources into a global model to achieve similar optimization results. This demonstrates the potential of federated learning for multi-domain collaboration, especially in scenarios where privacy protection is critical, to enable effective model updates and optimizations in different environments.

## V. CONCLUSION

In this paper, we explored the application of federated learning in cross-domain data privacy protection, focusing on how it can achieve efficient model training and optimization while ensuring data privacy. With the increasing prevalence of data sharing, particularly in sensitive fields such as healthcare, finance, and social media, data privacy security has become increasingly important. Traditional centralized learning methods typically require data to be collected and processed on a central server. While this approach can achieve good training results, the risk of privacy leakage is a major concern when handling cross-domain data. Federated learning, as an emerging distributed learning method, addresses this issue by keeping data local and only sharing model update parameters, effectively preventing data leakage while enabling knowledge sharing and collaboration among multiple participants.

Experimental results show that federated learning can maintain good training performance across data from different domains. By aggregating model updates from multiple data sources, the global model performs excellently in terms of accuracy, precision, recall, and F1-score. Notably, federated learning offers clear advantages over traditional methods in terms of privacy protection. Through simulating applications in healthcare, finance, and user data, we found that despite the heterogeneity of data sources, federated learning can effectively handle cross-domain data while ensuring privacy security.

Although federated learning has many advantages, it still has some problems in practice. Firstly, due to the heterogeneity of data distribution across clients, tackling the challenge of Non-IID data and stabilizing and optimizing the global model is an urgent problem in current federated learning research. Secondly, the efficiency of federated learning also needs further optimization. In scenarios with a large number of participants or large datasets, the communication and computational overhead can be substantial, potentially affecting the system's real-time performance. Therefore, future research could explore ways to improve the communication efficiency of federated learning, optimize the convergence speed of algorithms, and address how to handle more diverse data and tasks in cross-domain scenarios.